\documentclass{article} 
\usepackage{iclr2021_conference,times}

\usepackage{amsmath,amsfonts,bm}

\def\eqref#1{equation~\ref{#1}}

\def\1{\bm{1}}

\DeclareMathAlphabet{\mathsfit}{\encodingdefault}{\sfdefault}{m}{sl}
\SetMathAlphabet{\mathsfit}{bold}{\encodingdefault}{\sfdefault}{bx}{n}

\usepackage{hyperref}
\usepackage{url}

\title{Explainability for fair machine learning}

\author{%
\textbf{Tom Begley, ~~ Tobias Schwedes, ~~ Christopher Frye, ~~\&~~ Ilya Feige} \\[15pt]
Faculty, 54 Welbeck Street, London, UK
}

\usepackage[utf8]{inputenc} 
\usepackage[T1]{fontenc}    
\usepackage{hyperref}       
\usepackage{url}            
\usepackage{amsfonts}       
\usepackage{amsmath}
\usepackage{amsthm}
\usepackage{nicefrac}       
\usepackage{microtype}      
\usepackage{enumerate}
\usepackage{xcolor}

\usepackage{booktabs}
\usepackage{multirow}
\usepackage{multicol}
\usepackage{rotating}
\usepackage{array}

\newcommand{\eqn}[1]{\begin{align}#1\end{align}}

\DeclareRobustCommand{\Sec}[1]{Sec.~\ref{sec:#1}}

\DeclareRobustCommand{\App}[1]{App.~\ref{app:#1}}

\DeclareRobustCommand{\Tab}[1]{Table~\ref{tab:#1}}

\DeclareRobustCommand{\Fig}[1]{Fig.~\ref{fig:#1}}

\DeclareRobustCommand{\Eq}[1]{Eq.~(\ref{eq:#1})}

\def\cX{\mathcal{X}}
\newcommand{\bbE}{\mathbb{E}}

\usepackage{float}
\floatstyle{plaintop}
\restylefloat{table}

\usepackage{enumitem}

\usepackage{graphicx}
\usepackage{subcaption}
\usepackage{caption}

\usepackage{bm}

\usepackage{cleveref}
\graphicspath{ {./figures/} }

\newtheoremstyle{definition}
  {}   
  {}   
  {\itshape}  
  {0pt}       
  {\bfseries} 
  {}          
  {5pt plus 1pt minus 1pt} 
  {\thmname{#1}\thmnumber{ #2}.\thmnote{ \normalfont\uppercase{#3}}} 

\theoremstyle{definition}

\usepackage{color}
\definecolor{darkred}{rgb}{1.0,0.1,0.1}
\definecolor{darkgreen}{rgb}{0.1,0.7,0.1}
\definecolor{darkblue}{rgb}{0.1,0.1,1.0}
\definecolor{darkorange}{rgb}{1.0, 0.55, 0.0}

\iclrfinalcopy 
\begin{document}

\maketitle

\begin{abstract}

  As the decisions made or influenced by machine learning models increasingly impact our lives, it is crucial to detect, understand, and mitigate unfairness. But even simply determining what ``unfairness'' should mean in a given context is non-trivial: there are many competing definitions, and choosing between them often requires a deep understanding of the underlying task. It is thus tempting to use model explainability to gain insights into model fairness, however existing explainability tools do not reliably indicate whether a model is indeed fair. In this work we present a new approach to explaining fairness in machine learning, based on the Shapley value paradigm. Our fairness explanations attribute a model's overall unfairness to individual input features, even in cases where the model does not operate on sensitive attributes directly. Moreover, motivated by the linearity of Shapley explainability, we propose a meta algorithm for applying existing training-time fairness interventions, wherein one trains a perturbation to the original model, rather than a new model entirely. By explaining the original model, the perturbation, and the fair-corrected model, we gain insight into the accuracy-fairness trade-off that is being made by the intervention. We further show that this meta algorithm enjoys both flexibility and stability benefits with no loss in performance.

\end{abstract}

\section{Introduction}
\label{sec:intro}
Machine learning has repeatedly demonstrated astonishing predictive power due to its capacity to learn complex relationships from data.  However, it is well known that machine learning models risk perpetuating or even exacerbating unfair biases learnt from historical data \citep{barocas2016big, bolukbasi2016man, caliskan2017semantics, lum2016predict}. As such models are increasingly used for decisions that impact our lives, we are compelled to ensure those decisions are made fairly.

We consider fairness in the context of supervised learning, where the task is for a model $f$ to predict a target $y$ from features $x$ while avoiding unfair discrimination with respect to a protected attribute $a$ (e.g.~sex or race). We allow, but do not require, $a$ to be a component of $x$.

In the pursuit of training a fair model $f$, one encounters the immediate challenge of how fairness should be defined. There exist a wide variety of definitions of fairness --- some based on statistical measures, others on causal reasoning, some imposing constraints on group outcomes, others at the individual level --- and each notion is often incompatible with its alternatives \citep{berk2018fairness, corbett2017algorithmic, kleinberg2016inherent, lipton2018does, pleiss2017fairness}. Deciding which measure of fairness to impose on $f$ thus requires extensive contextual understanding and domain knowledge. Further still, one should understand the downstream consequences of a fairness intervention before imposing it on the model's decisions \citep{hu2019disparate, liu2018delayed}.

To help understand whether a model is making fair decisions, and choose an appropriate notion of fairness, one might be tempted to turn to model explainability techniques. Unfortunately, it has been shown that many standard explanation methods can be manipulated to suppress the reported importance of the protected attribute without substantially changing the output of the model \citep{dimanov2020you}. Consequently such explanations are poorly suited to assessing or quantifying unfairness.

In this work, we introduce new explainability methods for fairness based on the Shapley value framework for model explainability \citep{datta2016algorithmic, kononenko2010efficient, lipovetsky2001analysis, lundberg2017unified, vstrumbelj2014explaining}. We consider a broad set of widely applied group-fairness criteria and propose a unified approach to mitigating unfairness within any one of them. This set of fairness criteria includes both \emph{demographic parity} \citep{calders2009building, feldman2015certifying, kamiran2012data, zafar2017fairness_independence}, which calls for $f(x)$ to be unconditionally independent of $a$, and \emph{equalised odds} \citep{hardt2016equality}, which requires $f(x)$ to be independent of $a$ given $y$. It also includes \emph{conditional demographic parity} \citep{corbett2017algorithmic}, which calls for $f(x)$ to be independent of $a$ given a set of variables $\{v_i\}$. This generalisation of demographic parity introduces \emph{resolving variables}, $v_i$, with respect to which it is not considered unfair to discriminate. We show that for each of these definitions it is possible to choose Shapley value functions which capture the overall unfairness in the model, and attribute it to individual features. We also show that because the fairness Shapley values collectively must sum to the chosen fairness metric, we cannot hide unfairness by manipulating the explanations of individual features, thereby overcoming the problems with accuracy-based explanations observed by \cite{dimanov2020you}.

Motivated by the attractive linearity properties of Shapley value explanations, we also introduce a meta algorithm for training a fair model. Rather than learning a fair model directly, we propose learning a fairness-imposing perturbation to an existing unfair model using existing training-time algorithms for fairness. We show that this approach gives new perspectives helpful for understanding fairness, benefits from greater flexibility due to model-agnosticism, and enjoys improved stability, all while maintaining the performance of training-time interventions.

\section{Explainable fairness}
\label{sec:framework}
In this section we give an overview of the Shapley value paradigm for machine learning explainability, and show how it can be adapted to explain fairness. Motivated by the axiomatic properties of Shapley values, we also introduce a meta algorithm for applying training-time fairness algorithms to a perturbation rather than a fresh model, giving us multiple perspectives on fairness.

\subsection{Adapting explainabilty to fairness}
\label{sec:expl_details}

Fairness in decision making -- automated or not -- is a subtle topic. Choosing an appropriate definition of fairness requires both context and domain knowledge. We might be tempted to improve our understanding of the problem using model explainability methods. However \cite{dimanov2020you} show that such methods are poorly suited for understanding fairness. In particular we should not try to quantify unfairness by looking at the feature importance of the protected attribute, as such measures can be easily manipulated. Part of the problem is that most explainability methods attempt to determine which features are important contibutors to the model's predictive power. The importance of the protected attribute to accuracy may be related to, but is not sufficient to understand, fairness. Ideally we would have access to explainability methods that can interrogate fairness in the model and complement the explanation of accuracy.

Toward this end, we develop new fairness explanations within the Shapley value paradigm, which is widely used as a model-agnostic and theoretically principled approach to model explainability \citep{datta2016algorithmic, kononenko2010efficient, lipovetsky2001analysis, lundberg2017unified, vstrumbelj2014explaining}. We will first review the application of Shapley values to explaining model accuracy, then show how this can be adapted to explaining model unfairness. See \cite{frye2019asymmetric} for a detailed analysis of the axiomatic foundations of Shapley values in the context of model explainability. For clarity of exposition we assume a binary label $y$ so that $f(x)$ represents the probability that $y=1$. See \App{gen-explain} for straight-forward generalisations to multiclass classification.

\subsubsection*{Explaining model accuracy}
\label{sec:shapley-performance}

Shapley values provide a method from cooperative game theory to attribute value to the individual players on a team $N=\{1,\ldots,n\}$ \citep{shapley1953value}. If the team earns a total value $v(N)$, the Shapley value $\phi_v(i)$ attributes a portion to player $i$ according to:
\eqn{
    \label{eq:shapley}
    \phi_v(i) = \sum_{S \subseteq N \setminus \{i\}}
    \frac{|S|! \, (n-|S|-1)!}{n!} \,
    \big[ v(S \cup \{i\}) - v(S) \big]
}
Here $v(S)$ is the value function indicating the value a coalition $S$ of players receives when playing on their own. The Shapley value $\phi_v(i)$ is thus the marginal contribution that player $i$ makes upon joining a coalition, averaged over all coalitions and all orders in which those coalitions can form. 

To apply Shapley values to model explainability, one interprets the input features as the players of the cooperative game and defines an appropriate value function (e.g.~the model's output) to insert into \Eq{shapley}. We introduce
\eqn{
    f_y(x) = (1 - y)(1 - f(x)) + y \, f(x)
}
which denotes the predicted probability that $x$ belongs to class $y$. To explain this prediction $f_y(x)$, it is conventional to define a value function by marginalising over out-of-coalition features:
\eqn{
    \label{eq:value-function}
    v_{f_y(x)} (S) = \bbE_{p(x')} \big[f_y(x_S \sqcup x'_{N\setminus S})\big]
}
where $x_S$ is the set of feature values with indices in $S$ and $\cdot \sqcup \cdot$ denotes splicing disjoint sets of features to create a new data point, and where $p(x')$ represents the data distribution.\footnote{
    \label{fn:on-manifold}
    A more natural value function would marginalise over the conditional data distribution $p(x'_{N\setminus S} | x_S)$ as discussed in \cite{aas2019explaining} and \cite{frye2020shapley}. \Eq{value-function} is used here for clarity of exposition as it is more commonly found in the literature.
}
One computes local Shapley values $\phi_{f_y(x)}(i)$ by inserting $v_{f_y(x)}$ into \Eq{shapley}. These can be aggregated to obtain a global explanation of the model's performance that maintains the underlying Shapley axioms:
\eqn{
    \label{eq:global-shapley}
    \Phi_f(i) = \bbE_{p(x,y)} \big[
        \phi_{f_y(x)}(i)
    \big]
}
where $p(x, y)$ is the joint distribution from which the labelled data is sampled. Aggregating global Shapley values in this way provides the desirable property that
\eqn{
    \sum_i \Phi_f(i) = \mathbb E_{p(x, y)} \big[ f_y(x) \big] - \mathbb E_{p(x')p(y)} \big[ f_y(x') \big]
}
The first term on the right-hand side is the expected accuracy for a model which samples a predicted label according to the predicted probability. The second is an offset term corresponding to the accuracy that is not attributable to any of the features and is related to the class balance. We remark that randomised classifiers, such as those used by the reductions approach of \cite{agarwal2018reductions}, sample predicted labels in this way so expected accuracy coincides with commonly used deterministic accuracy. More generally, the expected accuracy captures the confidence with which the classifier makes predictions as well as whether the predictions are correct.

\subsubsection*{Explaining model fairness}

To explain fairness in a model's decisions, we define a new value function that captures this effect, then proceed as in \Sec{shapley-performance}. We focus on demographic parity here under the assumption that $a$ is binary. See \App{gen-explain} for straightforward generalisations to other notions of fairness, as well as to multiclass classification. To begin, let
\eqn{
\label{eq:fair-shapley-base}
    g_a(x) = f(x) \cdot \frac{(-1)^a}{p(a)}
}
in which the sign of $g_a(x)$ is controlled by the true value of the sensitive attribute $a$, whether or not the $a$-component in $x$ is altered. The value function on coalitions is defined through marginalisation:\footnote{This marginalisation could also be performed conditionally, as discussed in footnote \ref{fn:on-manifold}.}
\eqn{
\label{eq:fair-shapley-value-function}
    v_{g_a(x)} (S) = \mathbb{E}_{p(x')} \big[
        g_a(x_S \sqcup x'_{N\setminus S})
    \big]
}
Next, we insert $v_{g_a(x)}$ into \Eq{shapley} to obtain local Shapley values $\phi_{g_a(x)}(i)$. Finally, we obtain the corresponding global Shapley values through aggregation:
\eqn{
    \label{eq:global-fairness-shapley}
    \Phi_{g}(i) = \bbE_{p(x, a)}\big[ \phi_{g_a(x)}(i) \big]
}
where $p(x, a)$ is the joint distribution of features and protected attribute from which the data is sampled. The definitions above are motivated by the resulting sum rule for the global Shapley values:
\eqn{
    \label{eq:global-fairness-shapley-sum}
    \sum_i \Phi_{g}(i)
    = \int dx \, p(x|a=0) \, f(x) - \int dx \, p(x|a=1) \, f(x)
}
This is the expected demographic parity difference for a model which samples a predicted label according to the predicted probability. As discussed for the accuracy Shapley values, when we use randomised classifiers this expected demographic parity difference is the same as the regular demographic parity difference. The global Shapley values $\Phi_g(i)$ can thus be interpreted as each feature's marginal contribution to the overall demographic disparity in the model. The local Shapley values are less relevant, as the group notions of fairness we are interested are defined in terms of aggregate statistics over each group, and local Shapley values only explain individual predictions.

\subsection{Learning corrective perturbations}
\label{sec:corr_details}

The linearity axiom of the Shapley values guarantees that the fairness Shapley values of a linear ensemble of models are the corresponding linear combination of Shapley values of the underlying models. Motivated by this we consider the problem of learning an additive perturbation to an existing model in order to impose fairness. That is, given $f$ we would like to learn a parametric perturbation $\delta_\theta$ such that
\eqn{
    \label{eq:perturbed-model}
    f_\theta = f + \delta_\theta
}
is fair. We would then gain three different perspectives from the explanations of each component: the original sources of unfairness from $f$, the correction / trade-off that was made from $\delta_\theta$, and how corrected model makes fair predictions from $f_\theta$. Moreover, the explanations are mutually consistent.

We continue to focus on binary classification here, see \App{gen-pert-corr} for a generalisation to the multiclass classification case. Since we require that the output of the corrected model should also be a valid probability, we define $\delta_\theta$ via an auxiliary perturbation $\tilde \delta_\theta$ on the logit scale:
\eqn{
    \label{eq:logit-perturbed-model}
    \delta_\theta(f(x), x, a) = \sigma\Big(
        \sigma^{-1}\big( f(x) \big) + \tilde\delta_\theta\big(f(x), x, a\big)
    \Big) - f(x).
}
where $\sigma(x) = (1+\exp(-x))^{-1}$. We can apply a training-time fairness algorithm of our choice to $f_\theta$ in order to learn the parameters of the auxiliary perturbation $\tilde\delta_\theta$. In \Sec{results} we show results for the algorithms of \cite{agarwal2018reductions} and \cite{zhang2018mitigating} applied to the perturbed model.

Note that while in \Eq{logit-perturbed-model} we write $\delta_\theta$ as a function of $x$, $a$ and $f(x)$, it is possible to use only a subset of these inputs. If we only gave $\delta_\theta$ access to $a$ and $f(x)$ then this becomes a true post-processing algorithm. Alternatively we may require a fair algorithm that does not have access to the protected attribute at inference time, in which case we could omit $a$. We would then need to select a training-time algorithm that makes predictions without access to the protected attribute such as the reductions approach of \cite{agarwal2018reductions}.

\section{Results}
\label{sec:results}
In this section we showcase the fairness Shapley values introduced in \Sec{framework}. We further show that the models corrected through learning a perturbation suffer no loss in performance when compared to their unconstrained analogues, and enjoy increased flexibility and stability in addition to the explainability benefits. Our experiments are conducted on the Adult dataset from the UCI Machine Learning Repository \citep{Dua:2019}, where the task is to predict whether an individual earns more than \$50K per year based on their demographics, and the COMPAS recidivism dataset \citep{larson2016we}, where the task is to predict recidivism risk based on demographics. We use sex as the protected attribute in our experiments on Adult, and race in our experiments on COMPAS. Full details for all experiments are in \App{experimental-setup}.

\subsection{Explainability}
\label{sec:results-explainability}

\begin{figure}[!t]
    \centering
    \includegraphics[width=1.0\columnwidth]{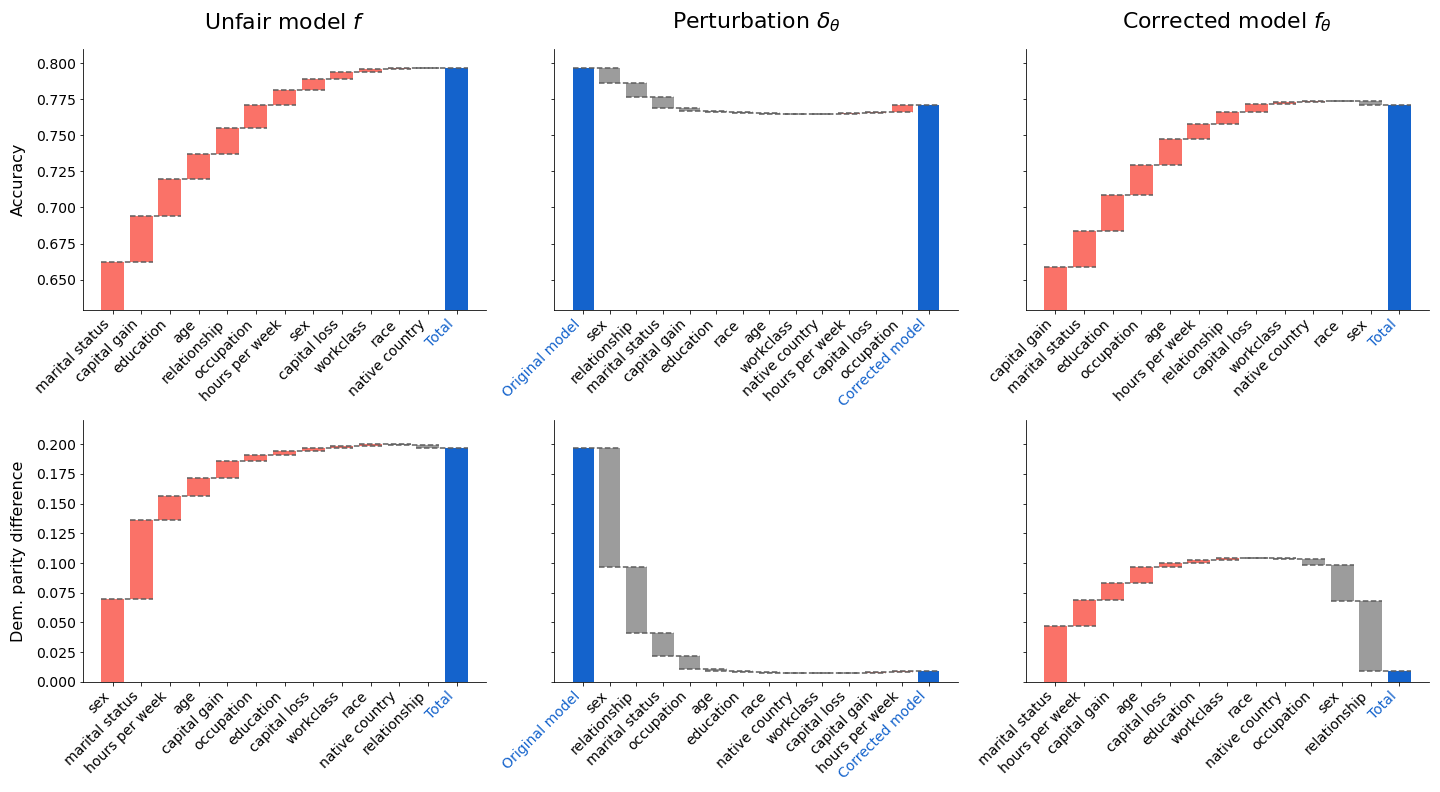}
    \caption{Explaining accuracy and unfairness (demographic parity) using Shapley values.}
    \label{fig:explainability}
    \vspace{-0.15cm}
\end{figure}

To test the effectiveness of the fairness Shapley values, we correct a model trained on the Adult dataset by applying the adversarial approach of \cite{zhang2018mitigating} to a perturbation of the baseline as described in \Sec{framework}. From left to right \Fig{explainability} shows the accuracy and fairness global Shapley values for the original model, perturbation, and corrected model respectively. We use waterfall plots to emphasise that they sum to the expected model accuracy or expected demographic parity difference respectively. The offset term for the accuracy Shapley values described in \Sec{framework} is used to offset the $y$-axis in the top row, fairness Shapley values have no offset term.

The accuracy Shapley values tell us which features contribute to the model accuracy, while the fairness Shapley values tell us which features contribute to the demographic parity difference. Comparing the two can give us useful insights into the legitimacy of a feature. For example we see that \texttt{marital-status} is the largest individual contributor to accuracy for the unfair model, but also makes a major contribution to unfairness. Weighing up the effects of each feature on both accuracy and fairness helps us determine whether it is legitimate to include them in the model.

We gain different insights into the problem from each of the three models. The explanation of the original-unfair model shows us why an unconstrained model might learn to make predictions in an unfair way, the explanation of the corrected-fair model shows us how it is able to make fair predictions, and the explanation of the perturbation captures the trade-off that has been made to impose fairness.

For example, in \Fig{explainability} we see that the unfair model uses all features to improve accuracy, as we would expect from a model that has been trained only to maximise accuracy. The accuracy Shapley values of the perturbation show that some predictive power from \texttt{sex} and its proxies \texttt{marital-status} and \texttt{relationship} has been sacrificed, but the fairness Shapley values show that as a result these three features were able to be used to heavily reduce the demographic parity difference. The fairness Shapley values for the corrected model show that the effects of \texttt{marital-status} and other features are almost perfectly counter-balanced by \texttt{sex} and \texttt{relationship}.

The complementary fairness and accuracy explanations, and the explanations offered from the three models have the potential to be a powerful tool for model development and fine-tuning of fairness.

\subsection{Robustness of fairness explanations}
\label{sec:results-robustness}

\begin{figure}[!t]
    \centering
    \includegraphics[width=0.85\columnwidth]{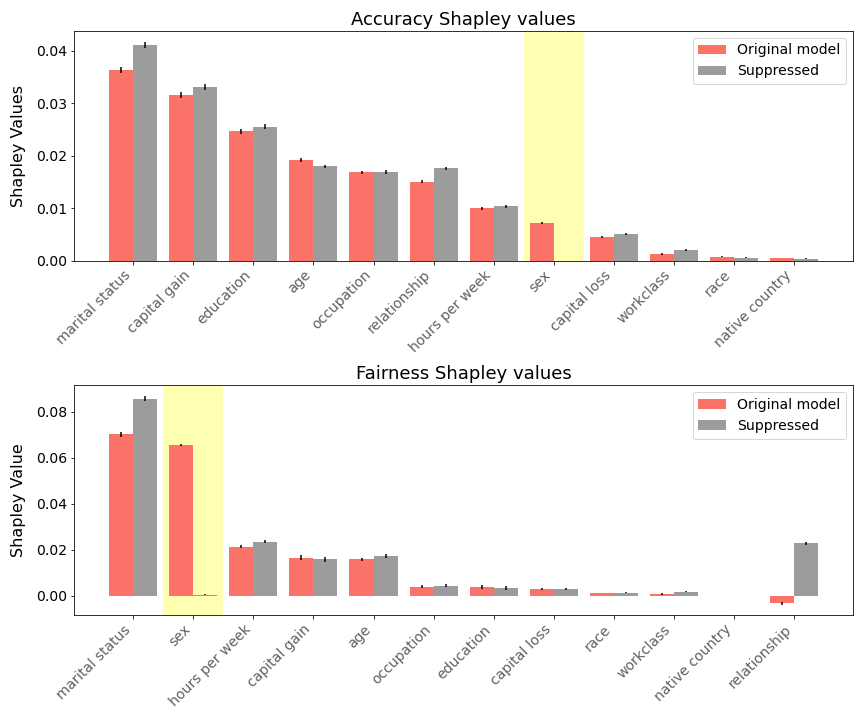}
    \caption{Suppressing the importance of the \texttt{sex} feature.}
    \label{fig:suppressed-explainability}
    \vspace{-0.15cm}
\end{figure}

\cite{dimanov2020you} show that existing explainability methods are poorly suited to answering questions about fairness. Specifically they show that by retraining a model with an additional penalty term corresponding to the influence the protected attribute has on the output, the explanation for that feature can be suppressed without substantially affecting the model predictions. Hence the feature importance of the protected attribute is a poor measure of fairness.

\Fig{suppressed-explainability} shows the accuracy and fairness Shapley values for the baseline model of the previous section, as well as the same model after retraining to suppress the influence of \texttt{sex} on the output. We see that retraining has completely eliminated \emph{functional} dependence of the output on \texttt{sex}, however we should not conclude that the model now satisfies demographic parity. Indeed the two models predict the same outcome on more than 98\% of the test data points, and demographic parity difference only improved fractionally, falling to 0.184 from 0.193.

The fairness Shapley values for \texttt{sex} has also been suppressed. This is a correct explanation, as the model output has no functional dependence on \texttt{sex}. However demographic parity requires \emph{statistical} independence which is not captured by the Shapley values. Indeed the fairness Shapley value of \texttt{sex} is zero, but the Shapley values for \texttt{marital-status} and \texttt{relationship} have gone up, showing that the model has simply switched focus to proxies of the protected attribute. We see the same effect in the accuracy Shapley values, however without the context of its impact on fairness it's hard to know whether the increased accuracy attributable to \texttt{relationship} is legitimate.

The fairness Shapley values additionally capture unfairness in the model through their sum, which equals the expected demographic parity difference. While we were able to manipulate individual Shapley values by encouraging the model to use proxies instead of the protected attribute, we are not able to reduce the sum of the fairness Shapley values without actually making the model satisfy demographic parity.

\subsection{Learnt perturbations}

In this section we investigate the properties of applying existing training time algorithms to a corrected model rather than a fresh model. We find that applying the reductions approach of \cite{agarwal2018reductions} and the adversarial approach of \cite{zhang2018mitigating} leads to no loss in performance, but we gain flexibility, stability and the benefit of being able to compare explanations between original model, perturbation and corrected model.

\subsubsection*{Performance}
\label{sec:performance}

\begin{table}[!t]
\scriptsize
\caption{Accuracy associated with decreasing demographic parity thresholds.}

\centering

\resizebox{.8\textwidth}{!}{

\begin{tabular}{  @{} c  @{} c @{} c @{} *6c @{\hspace{4mm}} c @{}}  \bottomrule
    \multirow{2}{*}{}
\hspace{0mm} & \multicolumn{1}{c}{}\hspace{25mm} & \multicolumn{8}{c}{\bf{Accuracy [$\%$] at demographic parity difference}} \\
    \cmidrule{3-10}

    & {\bf{Method}} &  & 0.1& 0.08& 0.06& 0.04& 0.02 & 0.01 & 0.005\\

\midrule

\multirow{5}{*}{\begin{turn}{90}\textbf{Adult}\end{turn} \hspace{2mm}}
    & {{Agarwal et al.}} &  & 84.71 & 84.32 & 83.94 & 83.82 & 83.29 & 83.29 & - \\
    & {{Agarwal et al. - perturbed}} &  & 84.69 & 84.43 & 83.82 & 83.82 & 83.35 & 83.23 & - \\
    & {{Zhang et al.}} &  & 84.65 & 84.18 & 84.06 & 83.58 & 83.18 & 83.15 & 83.15  \\
& {{Zhang et al. - perturbed }} &  & 84.74 & 84.48 & 83.78 & 83.61 & 83.14 & 82.99 & 82.96  \\
& {{Feldman et al. (post)}} &  & 84.69 & 84.35 & 84.12 & 83.67 & 83.32 & 83.30 & 83.01  \\

\midrule

\multirow{5}{*}{\begin{turn}{90}\textbf{COMPAS}\end{turn}\hspace{2mm}}
    & {{Agarwal et al.}}  & & 74.05 & 74.05 & 73.77 & 73.67 & 73.11 & 73.11 & 73.01  \\
    & {{Agarwal et al. - perturbed}} &  & 74.24 & 74.24 & 73.86 & 73.86 & 73.20 & 72.73 & 72.73  \\
    & {{Zhang et al.}} &  &  75.19 & 75.19 & 75.19 & 74.62 & 74.15 & 74.15 & 74.15  \\
& {{Zhang et al. - perturbed}} &  &	74.24 & 74.24 & 74.24 & 73.30 & 73.30 & 73.20 & 72.73  \\
& {{Feldman et al. (post)}} &  & 74.81 & 74.81 & 74.81 & 74.24 & 74.24 & 73.20 & 72.35  \\
\bottomrule
\end{tabular}
\label{tab:results_independence}
}
\end{table}

\begin{table}[!t]
\scriptsize
\caption{Accuracy associated with decreasing equalised odds thresholds.}

\centering

\resizebox{.8\textwidth}{!}{

\begin{tabular}{  @{} c  @{} c @{} c @{} *6c @{\hspace{4mm}} c @{}}  \bottomrule
    \multirow{2}{*}{}
\hspace{0mm} & \multicolumn{1}{c}{}\hspace{25mm} & \multicolumn{8}{c}{\bf{Accuracy [$\%$] at equalised odds difference}} \\
    \cmidrule{3-10}

    & {\bf{Method}} &  & 0.1& 0.08& 0.06& 0.04& 0.02 & 0.01 & 0.005\\

\midrule

\multirow{5}{*}{\begin{turn}{90}\textbf{Adult}\end{turn} \hspace{2mm}}
    & {{Agarwal et al.}} &  & 85.32 & 85.32 & 85.13 & 84.30 & 84.18 & - & - \\
    & {{Agarwal et al. - perturbed}} &  & 85.43 & 85.43 & 85.31 & 84.34 & 84.21 & - & - \\
    & {{Zhang et al.}} &  & 85.13 & 85.04 & 85.04 & 84.86 & 84.33 & 75.43 & 75.43  \\
& {{Zhang et al. - perturbed }} &  & 85.26 & 85.11 & 85.06 & 84.97 & 84.23 & 83.53 & -  \\
& {{Hardt et al.}} &  & 82.77 & 82.77 & 82.77 & 82.77 & 82.77 & 82.77 & 82.77  \\

\midrule

\multirow{5}{*}{\begin{turn}{90}\textbf{COMPAS}\end{turn}\hspace{2mm}}
    & {{Agarwal et al.}}  & & 75.19 & 75.19 & 74.05 & 74.05 & 73.86 & 73.39 & 73.39  \\
    & {{Agarwal et al. - perturbed}} &  & 74.43 & 74.43 & 74.43 & 73.86 & 73.39 & 73.39 & 73.39  \\
    & {{Zhang et al.}} &  &  74.62 & 74.62 & 74.62 & 74.62 & 74.62 & 73.48 & 53.12  \\
& {{Zhang et al. - perturbed}} &  &	74.34 & 74.34 & 74.34 & 74.34 & 73.48 & 72.44 & 72.44  \\
& {{Hardt et al.}} &  & 71.31 & 71.31 & 71.31 & 70.45 & 68.75 & - & -  \\
\bottomrule
\end{tabular}
\label{tab:results_separation}
}
\end{table}

We proposed the learnt perturbations in \Sec{framework} primarily due to their explainability benefits. It is important however that we verify that in doing so we have not compromised the performance of the algorithms. In order to do so we use the adversarial approach of \cite{zhang2018mitigating} and the reductions approach of \cite{agarwal2018reductions} to impose both demographic parity and equalised odds on the Adult and COMPAS datasets. In both cases we apply the algorithms to fresh models, and also to perturbed models as described in \Sec{framework}. We also compare to simpler post-processing approaches of \cite{feldman2015certifying} for demographic parity, and \cite{hardt2016equality} for equalised odds.

\Tab{results_independence} shows the results of our demographic parity experiments, and \Tab{results_separation} shows the results of the equalised odds experiments. We report for each method the highest observed test-set accuracy where the corresponding fairness metric did not exceed a specified threshold. All methods are seen to lose accuracy as the fairness requirement becomes more stringent, some methods performing better on one task than the other. The perturbed models perform competitively on both datasets and for both definitions of fairness, showing no significant reduction in accuracy at each fairness threshold over their training-time counterparts. For demographic parity we note that while the simple post-processing method of \citet{feldman2015certifying} performs competitively with the more sophisticated algorithms, it requires access to the protected attribute at inference time, which is not true of the approaches of \citet{agarwal2018reductions} and \citet{zhang2018mitigating} which only require access to the protected attribute during training.

\subsubsection*{Flexibility}

The meta algorithm presented in \Sec{framework} offers flexibility benefits that are typical of post-processing algorithms while inheriting the performance benefits of the training-time algorithms it utilises to impose fairness. Specifically, we can be fully model-agnostic with respect to the original model, as any model structure or access requirements apply only to the perturbation, and not the original model. Indeed we could learn a perturbation to a model that is only available for inference over a network, or correct a rules based classifier that does not lend itself to conventional optimisation techniques. Furthermore, if the original model is complex, we have the option of training a lightweight perturbation to the complex model, and may not need to rerun an expensive training procedure.

\subsubsection*{Stability}

\begin{figure}[t!]
    \centering
    \includegraphics[width=1.0\columnwidth]{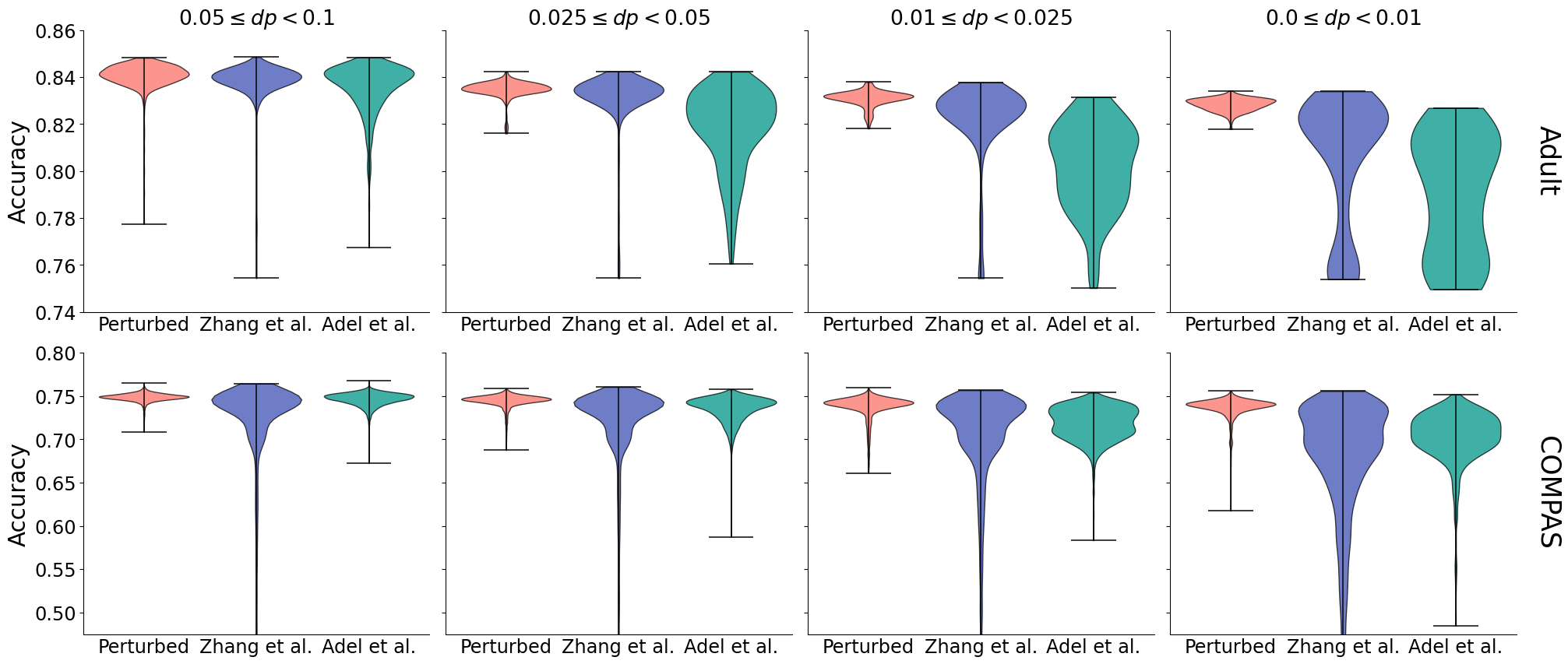}
    \caption{Accuracy violin plots of experimental outcomes binned by achieved level of fairness.}
    \label{fig:orchestra}
\end{figure}

Since accuracy and fairness objectives are typically in tension, some optimisation methods exhibit training-time instability where the classifier does not converge to a local minimum, but instead jumps between them. This is particularly true of the adversarial method of \cite{zhang2018mitigating}, where the tension between accuracy and fairness manifests itself as the tension between the model and the adversary. Indeed training-time stability is a generic problem for adversarial methods. \cite{adel2019one} introduce a variant of adversarial debiasing that attempts to address this instability.

Our perturbation-based approach combats the instability introduced by this tension between competing objectives: the fair-corrected model remains anchored near the original-unfair model, which has already found a local-accuracy optimum. This is also evident in \Tab{results_separation} where we observe that as we make the fairness threshold more stringent, classifiers trained with the method of \cite{zhang2018mitigating} collapse to a constant predictor, predicting according to class balance, whereas the perturbed model, optimised in the same way, does not exhibit the same behaviour and is able to learn a non-trivial classifier with low equalised odds difference.

We illustrate the improved stability by comparing the application of adversarial training to a perturbation, and the original approaches of \cite{zhang2018mitigating} and \cite{adel2019one}. For each approach we train multiple models with a variety of hyperparameter choices. The resulting outcomes show the spread of model behaviour during training. We visualise the results in \Fig{orchestra}, which shows the distribution of accuracy outcomes in every experiment, binned by the achieved level of fairness. The plots show that while the optimal accuracy in each bin is similar across the three methods, our perturbative approach generally has less variance and higher mean accuracy across the range of hyperparameters considered. We found similar results when imposing equalised odds (see \App{equalised_odds}) thus showing a clear training-time stability advantage of our perturbation-based approach across multiple notions of fairness.

\section{Conclusions}
\label{sec:conclusions}
In this work we introduced a new approach to explaining fairness in machine learning, based on the Shapley value paradigm. The fairness Shapley values are able to attribute unfairness in a model's predictions to individual features, and complement the already well-used accuracy Shapley values. We also showed that because these explanations directly capture unfairness through their sum, they are robust to manipulation. Additionally, motivated by the linearity properties enjoyed by Shapley values, we introduced a meta algorithm for learning a fairness-imposing perturbation to an unfair model using existing training-time algorithms for fairness. This gives us new perspectives from the corresponding explanations: we are able to explain the unfair model to understand why model predictions might be at risk of being unfair, the corrected model to understand how it is able to make fair predictions, and the perturbation to understand the trade-off that has been made between the two models. Moreover, we showed that training a perturbation of this kind results in no loss in performance, but increases flexibility and can improve the training-time stability by anchoring the corrected model to a local-accuracy optimum. We believe that our approach to explaining fairness will be highly valuable to practitioners, and we hope they will contribute to the development of fairer models, and a greater understanding of unfairness in different machine learning problems.

\bibliography{references}
\bibliographystyle{iclr2021_conference}

\newpage

\appendix
\section{Appendix}
\subsection{Perturbing a multiclass classifier}
\label{app:gen-pert-corr}
In \Sec{framework} we introduced a meta algorithm for correcting an unfair model by learning a perturbation assuming a binary classification task, and remarked that it was straight-forward to generalise to multiclass classification. In this section we give the details.

We assume labels $y \in \{1,\dots, k\}$ and a model $f: \cX \rightarrow [0, 1]^k$ that maps each datapoint $x$ to a categorical probability distribution over the $k$ classes. We denote by $f_y(x)$ the probability that $x$ belongs to class $y$. By assumption, we have $\sum_{y=1}^k f_y(x) = 1$ for all $x$.

As in \Eq{perturbed-model}, we seek to introduce a perturbation $\delta_\theta$ to produce a fair model $f_\theta$. As in \Eq{logit-perturbed-model} we will define $\delta_\theta$ via an auxiliary perturbation on an unconstrained scale in order that we maintain valid probabilities. It is conventional to use the softmax function to convert a $k$-vector $z$ of logits to a categorical probability distribution
\eqn{
    \mathrm{softmax}(z)_i = \frac{\exp(z_i)}{\sum_j \exp(z_j)}
}
The softmax function is invariant under addition of a constant to all arguments. As a result, there is a translation invariance in the space of perturbations, which is not desirable for efficient optimisation. We circumvent this issue by pinning the first logit to zero. Specifically we define
\eqn{
    l(s)_i := \log(s_i) - \log(s_1) \quad i = 1, \dots, k
}
which as one easily verifies is a right-inverse for $\mathrm{softmax}$ on the space of $k$-simplices, i.e. $\mathrm{softmax}(l(s)) = s$ for any $k$-simplex $s$. With this in hand we define the perturbation analogously to \Eq{logit-perturbed-model} as follows
\eqn{
    \label{eq:multiclass-perturbation}
    \delta_\theta(f(x), x, a) = \mathrm{softmax}(l(f(x)) + \tilde\delta_\theta(f(x), x, a)) - f(x)
}
where we require that $(\tilde\delta_\theta)_1 \equiv 0$. Thus we perturb only $k-1$ of the logits for a $k$-category classification problem. With the corrected model defined, we proceed precisely as before, applying a training-time algorithm to $f_\theta$ in order to learn the parameters of $\delta_\theta$.

We note that this setup is easily seen to be equivalent to the formulation in \Eq{logit-perturbed-model} in terms of the sigmoid and logit functions when $k=2$. Indeed in that case we have $l(z)_1 = 0$ for all $z$, and $l(z)_2 = \log(z_2) - \log(z_1) = \log(z_2) - \log(1 - z_2) = \sigma^{-1}(z_2)$. Moreover by choice of $\tilde \delta_\theta$ the first component of the argument of $\mathrm{softmax}$ on the right hand side of \Eq{multiclass-perturbation} is $0$, and we have $\mathrm{softmax}((0, z))_2 = e^z (1 + e^z)^{-1} = \sigma(z)$.

\subsection{Explainability for other fairness notions}
\label{app:gen-explain}
To generalise the fairness explanations to other notions of fairness, we need only to adjust the weighting in \Eq{fair-shapley-base} and the distribution in \Eq{global-fairness-shapley} over which we aggregate to produce global Shapley values. We present the details below. As in \App{gen-pert-corr} we assume a multiclass classification task where $f_y(x)$ denotes the probability that $x$ belongs to class $y$.

\subsubsection*{Equalised odds}

The fairness Shapley values presented in \Sec{framework} can also be used to determine each feature's marginal contribution to equality of odds. Specifically we modify \Eq{fair-shapley-base} to create new functions
\begin{equation}
    g_{a, y}(x) =
    \begin{cases}
        \frac{ f_y(x) }{ P(a = 0 | y) } \quad & \text{if } a = 0 \\
        -\frac{ f_y(x) }{ P(a \neq 0 | y) } \quad & \text{if } a \neq 0
    \end{cases}
\end{equation}
for each $y$. As before we then create a value function by marginalising over out-of-coalition features
\begin{equation}
    v_{g_{a,y}(x)} (S) = \mathbb{E}_{p(x')} \left[ g_{a,y}(x_S \sqcup x'_{\bar S}) \right].
\end{equation}
Shapley values are then calculated using \Eq{shapley} to average over all possible coalitions. To compute global Shapley values we aggregate using \Eq{global-fairness-shapley}, but replacing $p(x, a)$ with $p(x, a | y)$, from which we obtain the following analogue of \Eq{global-fairness-shapley-sum}
\begin{equation}
    \sum_i \Phi_{g_y}(i)
    = \int dx \, p(x|y, a=0) \, f_y(x) - \int dx \, p(x|y, a\neq 0) \, f_y(x)
\end{equation}
This corresponds to the (signed) difference in expected sensitivity to label $y$ between the protected group $a=0$ and all other groups for a classifier that samples predicted labels according to the predicted probabilities. In the case of binary classification, we can interpret each integral as the expected true positive and true negative rates when $y = 1$ and $y = 0$ respectively.

For equality of odds, we proceed exactly as above, but we are concerned only with the case where $y = 1$.

\subsubsection*{Conditional demographic parity}
We proceed similarly to obtain explanations of fairness for conditional demographic parity by first defining
\begin{equation}
    g_{a, v_1, \dots, v_n}(x) =
    \begin{cases}
        \frac{ f_y(x) }{ P(a = 0 | v_1, \dots, v_n) } \quad & \text{if } a = 0 \\
        -\frac{ f_y(x) }{ P(a \neq 0 | v_1, \dots, v_n) } \quad & \text{if } a \neq 0
    \end{cases}
\end{equation}
for each possible combination of resolving variables $v_1, \dots, v_n$ and value of $y$, then creating a value function by marginalising over out-of-coalition features
\begin{equation}
    v_{g_{a, y, v_1, \dots, v_n}(x)} (S) = \mathbb{E}_{p(x')} \left[ g_{a,y,v_1, \dots, v_n}(x_S \sqcup x'_{\bar S}) \right].
\end{equation}
Shapley values are then calculated using \Eq{shapley} to average over all possible coalitions. To compute global Shapley values we aggregate using \Eq{global-fairness-shapley}, but replacing $p(x, a)$ with $p(x, a | v_1, \dots, v_n)$, from which we obtain the following analogue of \Eq{global-fairness-shapley-sum}
\begin{equation}
    \sum_i \Phi_{g_{y, v_1, \dots, v_n}}(i)
    = \int dx \, p(x|a = 0, v_1, \dots, v_n) \, f_y(x) - \int dx \, p(x|a\neq 0, v_1, \dots, v_n) \, f_y(x)
\end{equation}
which represents the signed expected conditional demographic parity difference for the protected groups $a = 0$ and $a \neq 0$ (conditional on $v_1, \dots, v_n$).

\subsection{Experimental setup}
\label{app:experimental-setup}

In this section we give details on the datasets and experimental results presented in \Sec{results}. All experiments were written in Python, we used TensorFlow to specify and train the models \citep{abadi2016tensorflow} and ran all of our experiments on AWS EC2 instances.

\subsubsection*{Datasets}
\subsubsubsection{\bf Adult}

The Adult dataset \citep{Dua:2019} contains 48842 rows with 14 features, and is pre-split into 66.7\% training and 33.3\% testing data. We drop the \texttt{fnlwgt} feature as it represents weights used in the original census application not relevant to the task at hand, as well as the \texttt{education-num} feature since it is a different representation of the existing \texttt{education} feature. This leads to 12 features, namely: \texttt{age}, \texttt{workclass}, \texttt{education}, \texttt{marital-status}, \texttt{occupation}, \texttt{relationship}, \texttt{race}, \texttt{sex}, \texttt{capital-gain}, \texttt{capital-loss}, \texttt{hours-per-week}, \texttt{native-country}. We drop all rows with missing values. When encoding the \texttt{native-country} feature we group into ``other'' all countries other than US and Mexico. We obtain a validation set by randomly splitting off 20\% of the remaining training data, leading to a final split of 53.3\% training, 33.3 \% testing and 13.3\% validation data from a total of 45,222 rows. We one-hot encode all categorical features, and standardise the continuous features by subtracting the train set mean and scaling by train set standard deviation.

\subsubsubsection{\bf COMPAS}

We follow the pre-processing steps used in \citep{larson2016we} applied to the original dataset\footnote{\url{https://github.com/propublica/compas-analysis/}} which contains 7214 rows and 52 features: We remove rows with charge dates not within 30 days from the arrest date to ensure we relate to the right offence, and only select individuals of African-American and Caucasian race. Some features contain information contained in others, or are not relevant such as defendants' names. We hence select only a small subset of semantically most relevant features, namely: \texttt{age}, \texttt{sex}, \texttt{race}, \texttt{c\_charge\_degree}, \texttt{priors\_count}, \texttt{c\_jail\_in}, \texttt{jail\_time\_out}, \texttt{juv\_fel\_count}, \texttt{juv\_misd\_count}, \texttt{juv\_other\_count}. The final dataset contains 5278 rows and 10 features, which we randomly split into 60\% training, 20\% validation and 20\% testing data. The categorical variables are one-hot encoded, while the remaining features are standardised by subtracting the train set mean and scaling by train set standard deviation.

\subsubsection*{Experiments}

We here give the details of experimental setup for each of the methods we consider in our experiments in \Sec{results}. As much as possible when comparing different methods we attempted to use similar model structures.

\subsubsubsection{\bf Explainability}

To produce the results in \Sec{results-explainability} we trained a feedforward neural network (one hidden layer, 50 hidden units, and ReLU activations) on the Adult dataset. We then introduced a second neural network with the same architecture to serve as the perturbation. The original model is frozen and we apply the algorithm of \cite{zhang2018mitigating} to train the permutation. We then calculate Shapley values for each of the original model, the perturbation and the corrected model by first sampling coalitions of features, then approximating \Eq{value-function} -- resp. \Eq{fair-shapley-value-function} -- via Monte Carlo approximation with the empirical data distribution.

\subsubsubsection{\bf Robustness}

To produce the results of \Sec{results-robustness} we start with the baseline model of the previous section. We then retrain it to suppress the importance of \texttt{sex}. We are motivated by, but deviate slightly from, the work of \cite{dimanov2020you}. Whereas they propose adding a penalty term corresponding to the gradient of the loss with respect to \texttt{sex}, we instead use a finite difference of the model output with respect to \texttt{sex} due to the discrete nature of the feature. Thus the modified loss becomes
\eqn{
    \frac 1 N \sum_{i=1}^N \mathcal{L}(f(x_i), y_i) + \alpha \Big|f(x_i~ |~ \text{do}(\text{sex}=1)) - f(x_i~ |~ \text{do}(\text{sex}=0))\Big|
}
where $\mathcal{L}$ is the original cross-entropy loss, $\alpha$ is a hyperparameter controlling the trade-off between optimising the accuracy and minimising the effect of \texttt{sex}, and $f(x_i \,|\, \text{do}(\text{sex}=j))$ denotes, via a slight abuse of notation, $f$ evaluated on the data point $x_i$ with the value for \texttt{sex} replaced with $j$. We train the baseline model for an additional 200 batches with $\alpha = 3$. The resulting model agrees with the baseline on over 98.5\% of the data, and has the same test set accuracy. We calculate Shapley values for the retrained model as before.

\subsubsubsection{\bf Performance}

For \Sec{performance} we trained a single-layer feedforward neural network on each dataset to serve as a baseline. On the Adult dataset we use 50 hidden units, on COMPAS we used 32. In both cases we gave the network ReLU activations.

We use the training-time algorithms of \cite{agarwal2018reductions} and \cite{zhang2018mitigating} to train additional neural networks with the same architecture while imposing demographic parity and equalised odds. We used the implementation of the reductions approach of Agarwal et al. from the \texttt{fairlearn} library of \cite{bird2020fairlearn}, with some superficial modifications to make it compatible with our TensorFlow models. We implemented the adversarial approach of \cite{zhang2018mitigating} ourselves. Our implementation runs for a fixed number of iterations, then restores the weights corresponding to the best validation set loss during the second half of training. All models are then evaluated on the unseen test set.

We reapplied both of these algorithms to the problem of learning a perturbation to the baseline model. Again the perturbations were single layer neural networks with the same architecture as the baseline.

Finally we consider two post-processing methods, those of \cite{feldman2015certifying} \footnote{In the original paper, \cite{feldman2015certifying} present their algorithm as a pre-processing method. The idea of instead applying it as a post-processing algorithm is not ours, having been previously observed by Hardt \url{https://mrtz.org/nips17/\#/41}. We weren't able to find any additional sources for this idea.} and \cite{hardt2016equality}, to impose demographic parity and equalised odds respectively. We use our own implementation of the former, and the AI Fairness 360 library to apply the latter \citep{bellamy2018ai}, each applied to the aforementioned baseline.

\subsubsubsection{\bf Stability}

To demonstrate instability during training of the adversarial methods, we compare the approaches of \cite{adel2019one}, and \cite{zhang2018mitigating}, in the latter case applied both as specified by the authors, and also to a perturbation as specified in \Sec{framework}. \Tab{grid_parameters} show the selection of hyperparameters over which we search. We chose them to cover a wide range of plausible experimental configurations. For each combination we additionally vary the weight of the discriminator term in the loss -- controlling the relative importance of the fairness objective compared to the accuracy objective. We run 5 experiments with each combination and record the results after a fixed number of iterations to construct the violin plots in \Sec{results}.

\begin{table}
\scriptsize
\caption{Values for hyperparameter grid used in our simulations}
\centering

\resizebox{.72\textwidth}{!}{

\begin{tabular}{llcc}
\bottomrule

&
& \multicolumn{2}{c}{\bf{Dataset}} \\
 \cmidrule{3-4}

& {\bf{Variables}} &   Adult &   COMPAS \\
\midrule

 \multirow{4}{*}{\begin{turn}{90}\textbf{Model}\end{turn}} &
 {Number of hidden layers} &
	$\{$2, 3$\}$ & $\{$0, 1, 2$\}$ \\
& {Width of hidden layers} &
	32 & $\{$ 16, 24$\}$ \\
& {Learning rate} &
	$\{$0.0001, 0.001$\}$& $\{$0.0001, 0.001$\}$ \\
& {Iterations} &
	$\{$2500, 5000$\}$ & $\{$2500, 5000$\}$ \\
\midrule

 \multirow{4}{*}{\begin{turn}{90}\textbf{\tiny Adversary}\end{turn}} &
 {Number of hidden layers} &
	$\{$2, 3$\}$ & $\{$0, 1, 2$\}$ \\
& {Width of hidden layers} &
	32& $\{$16, 24$\}$ \\
& {Learning rate} &
	0.01 & 0.01 \\
& {Iterations per model iteration} &
	$\{$1, 5$\}$& $\{1$, 5$\}$ \\
\midrule

& {Batch size} &
	512 & 128 \\
\bottomrule

\end{tabular}
\label{tab:grid_parameters}
}
\end{table}

\begin{table}

\scriptsize
\caption{Number of observations associated with intervals of increasing fairness}

\centering

\resizebox{.78\textwidth}{!}{

\begin{tabular}{@{} c @{} c  @{} c  @{} c @{} c @{} *9c @{\hspace{4mm}} c @{}}  \bottomrule
 \multicolumn{2}{c}{} &
 \multirow{2}{*}{}\hspace{0mm} & \multicolumn{1}{c}{}\hspace{19mm} & \multicolumn{5}{c}{\bf{Number of observations in unfairness interval $I=$}} \\
 \cmidrule{6-9}

& & & {\bf{Method}} &  & [0.05, 0.1)&   [0.025, 0.05)&   [0.01, 0.025)&  [0, 0.01)\\

\midrule

\multirow{6}{*}{\begin{turn}{90}\textbf{Demographic}\end{turn}} &
\multirow{6}{*}{\begin{turn}{90}\textbf{parity}\end{turn}} \hspace{1.5mm}
&\multirow{3}{*}{\begin{turn}{90}\textbf{Adult}\end{turn}}
& Perturbed &  &
	423& 201& 146& 103 \\
&& & Zhang et al. &  &
	550& 281& 168& 183 \\
&& & Adel et al. &  &
	606& 214& 117& 39 \\
\cmidrule{3-9}
&&
\multirow{3}{*}{\begin{turn}{90}\textbf{\tiny{COMPAS}}\end{turn}}
& Perturbed &  &
	4009& 2548& 1956& 1459 \\
&& & Zhang et al. &  &
	2776& 2771& 2526& 2892 \\
&& & Adel et al. &  &
	10015& 3157& 1482& 945 \\
\midrule
\multirow{6}{*}{\begin{turn}{90}\textbf{Equalised}\end{turn}} &
\multirow{6}{*}{\begin{turn}{90}\textbf{odds}\end{turn}} \hspace{1.5mm}
&\multirow{3}{*}{\begin{turn}{90}\textbf{Adult}\end{turn}}
& Perturbed &  &
	509& 1083& 231& 76 \\
&& & Zhang et al. &  &
	384& 1140& 278& 112 \\
&& & Adel et al. &  &
	1554& 1902& 334& 14 \\
\cmidrule{3-9}
&&
\multirow{3}{*}{\begin{turn}{90}\textbf{\tiny{COMPAS}}\end{turn}}
& Perturbed &  &
	3125& 2761& 2854& 3228 \\
&& & Zhang et al. &  &
	1986& 2381& 3437& 4087 \\
&& & Adel et al. &  &
	6421& 6827& 6330& 3756 \\

\bottomrule

\end{tabular}
\label{tab:results_observations}
}

\end{table}

\subsection{Stability results for equalised odds}
\label{app:equalised_odds}

\begin{figure}[!t]
    \centering
    \includegraphics[width=\columnwidth]{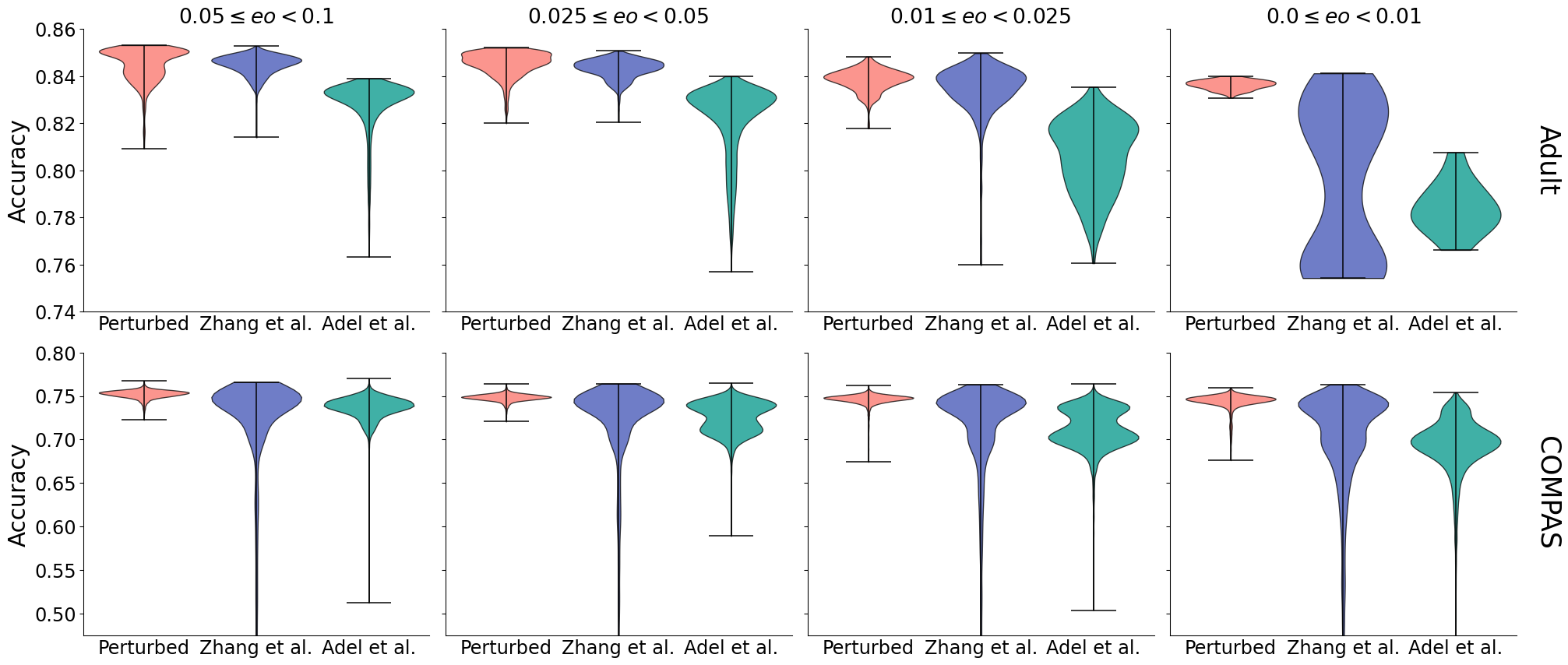}
    \caption{Accuracy violin plots of experimental outcomes binned by achieved level of fairness.}
    \label{fig:separation-orchestra}
\end{figure}

We also ran a grid search over hyperparameters with repeats to test the stability of the adversarial training of perturbations as compared to regular adversarial approaches. \Fig{separation-orchestra} shows the distribution of accuracy outcomes for each method on the two datasets we have been considering throughout, Adult and COMPAS. We see similar results to the demographic parity experiments in that the optimal accuracy in each fairness bin is comparable, but there is a much smaller spread of outcomes for the perturbation-based approach. This is because the perturbation remains anchored to a local accuracy optimum throughout, discouraging the model from jumping between local minima. \Tab{results_observations} shows the number of experimental results that fell in each bin for both the demographic parity, and equalised odds experiments.

\end{document}